\documentclass{article}
\usepackage{amssymb,amsfonts}
\usepackage{spconf,amsmath,graphicx,hyperref}
\usepackage{multirow}
\usepackage{xcolor}  
\usepackage{colortbl}
\usepackage{booktabs} 
\usepackage{threeparttable}

\title{CLIP-driven Zero-shot Learning with Ambiguous Labels}
\name{Jinfu Fan$^{1,\dagger}$, Jiangnan Li$^{1,\dagger}$, Xiaowen Yan$^{1}$, Xiaohui Zhong$^{1}$, Wenpeng Lu$^{2}$, Linqing Huang$^{3, *}$
\thanks{This research was supported by the National Natural Science Foundation of China (Nos. 62401309, 62376130, 62401362), the China Postdoctoral Science Foundation (No. 2025M771692), the Natural Science Foundation of Shandong Province and Qingdao (No. 2R2024QF119, No. 24-4-4-zrjj-89-jch), and open project (No. 2024PY030).}
\thanks{$\dagger$ Equal contribution. * The corresponding author.}
}
\vspace{-2mm}
\address{
$^{1}$ College of Computer Science and Technology, Qingdao University\\
$^{2}$ Key Laboratory of Computing Power Network and Information Security, Ministry of Education,\\ Shandong Computer Science Center,
Qilu University of Technology (Shandong Academy of Sciences)\\
$^{3}$ School of Computer Science, Shanghai JiaoTong University
}
\begin{document}
%
\maketitle
\begin{abstract}
Zero-shot learning (ZSL) aims to recognize unseen classes by leveraging semantic information from seen classes, but most existing methods assume accurate class labels for training instances. However, in real-world scenarios, noise and ambiguous labels can significantly reduce the performance of ZSL. To address this, we propose a new CLIP-driven partial label zero-shot learning (CLIP-PZSL) framework to handle label ambiguity. First, we use CLIP to extract instance and label features. Then, a semantic mining block fuses these features to extract discriminative label embeddings. We also introduce a partial zero-shot loss, which assigns weights to candidate labels based on their relevance to the instance and aligns instance and label embeddings to minimize semantic mismatch. As the training goes on, the ground-truth labels are progressively identified, and the refined labels and label embeddings in turn help improve the semantic alignment of instance and label features. Comprehensive experiments on several datasets demonstrate the advantage of CLIP-PZSL.
\end{abstract}

\vspace{-0.2cm}

\section{Introduction}
\label{sec:intro}

\vspace{-0.2cm}

Compared with traditional classification tasks, zero-shot learning (ZSL) \cite{F2,F27} transfers knowledge from seen classes to unseen classes via shared semantic information (e.g., attributes \cite{F3}, label vectors \cite{F4} or sentence descriptions \cite{F5}), enabling recognition without labeled instances of unseen classes. Despite its progress, most ZSL methods assume accurately labeled training data, whereas obtaining clean and complete labels is challenging, time-consuming, and not scalable. In practical scenarios, alternative solutions like crowdsourcing \cite{F7} and online queries \cite{F8} can reduce label costs but introduce noise and ambiguous labels, leading to ZSL overfitting to ambiguous labels and degraded performance.

To reduce the annotation workload, weakly supervised learning approaches like partial label learning (PLL) \cite{sun2024deep,F10,F11} have been explored, where each instance is linked to multiple candidate labels, but only one is correct. PLL relaxes labeling constraints, allowing ambiguous and noisy labels. However, PLL is limited to predicting only the seen classes, which restricts its capacity to recognize unseen classes.

To advance further, we propose a CLIP-driven partial label zero-shot learning (CLIP-PZSL) framework, which effectively alleviates the impact of ambiguity and noisy labels in training data and achieves the prediction of unseen classes by combining the strengths of ZSL and PLL. This setting is more complex, as ambiguous candidate labels mislead ZSL and harm generalization. The model must both handle ambiguity and maintain strong recognition for unseen classes. This raises a new challenge: How to learn semantic information from seen classes under the influence of ambiguous labels and effectively transform it into unseen classes for recognition.

To address these issues, we consider contrastive language-image pre-training (CLIP) \cite{F12}, which matches each input image to the most relevant text and shows strong zero-shot performance. However, ambiguous labels can cause semantic differences when embedding instance features into the label space, where distribution alignment is often beneficial \cite{Huang2025Incomplete}. For this problem, we propose CLIP-PZSL, which enhances the generalization performance of ZSL by instance-label alignment and label disambiguation. Specifically, CLIP encoders generate instance and label features, which are fused through a semantic mining block to adaptively extract discriminative features of different labels to learn a set of label embeddings for downstream disambiguation. This makes instance-label similarity measurable, enabling effective noisy labels detection and alignment of label and instance semantic vectors. Furthermore, we propose a robust partial zero-shot loss function to handle ambiguous labels in seen classes, which assigns weights to candidate labels based on instance-label relevance, guiding classifier training, and minimizing semantic mismatch. As the training goes on, the ground-truth labels are progressively identified, and the refined labels and class embeddings further improve the semantic alignment.

\begin{figure*}
\centerline{\includegraphics[width=0.98\textwidth]{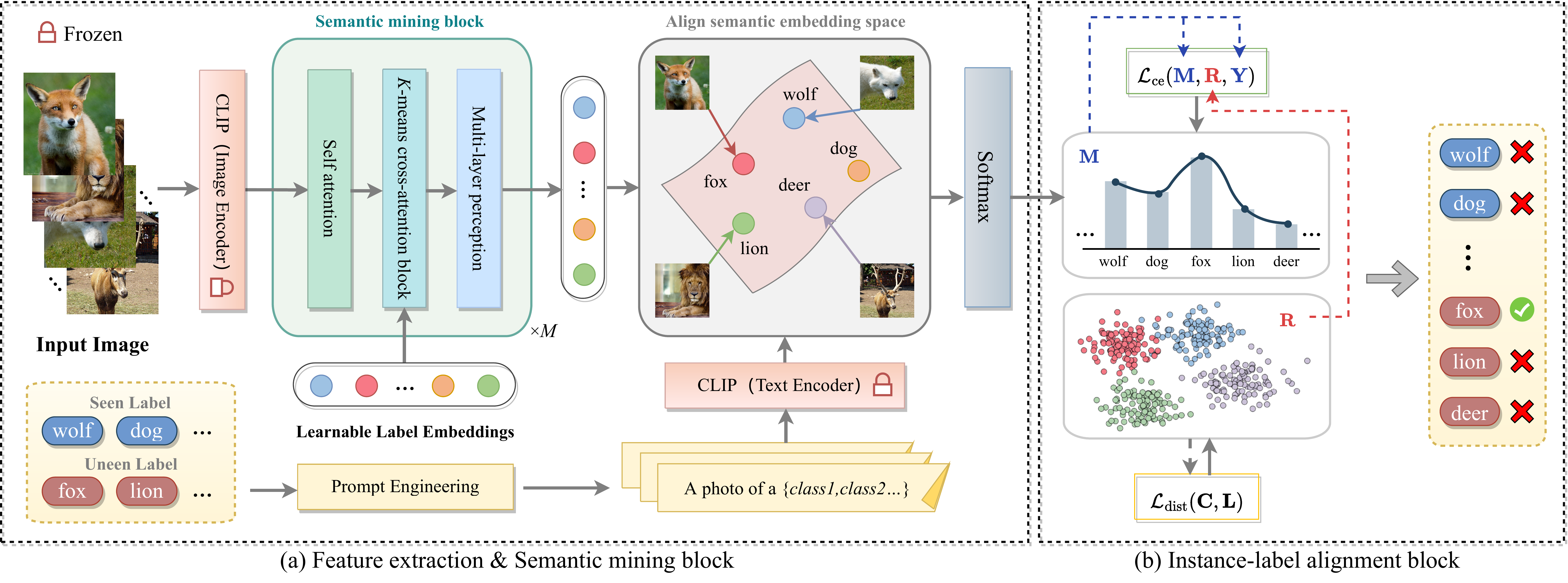}}
\vspace{-0.2cm}
\caption{Flowchart of CLIP-PZSL.}
\label{model}
\end{figure*}

Our key contributions are briefly summarized as follows: 

(1) To our knowledge, CLIP-PZSL is the first work for ZSL that effectively handles ambiguous labels in seen classes.

(2) We design a new semantic mining block from a clustering perspective to extract key information and align it with label embeddings for better noisy-label detection.

(3) We propose a robust partial zero-shot loss function for training CLIP-PZSL, which not only mitigates the impact of noisy labels but also aligns instance and label embeddings in the same dimension to minimize semantic mismatch.

\section{The Proposed Approach}
\label{sec:appr}

The problem of CLIP-PZSL is formulated as follows. Let $\mathcal{X}$ denote the instance space and $\mathcal{Y}_{s}=\left\{  y_{c}:c=1,\ldots,Q \right\}$ denote the output space with $Q$ class labels. The training dataset is $\mathcal{D}=\left\{ (\mathbf{x}_{i},S_{i}) |1\leq i\leq N \right\}$, where $\mathbf{x}_{i}\subseteq \mathcal{X}$ is a feature vector, $N$ is the number of instances, $S_{i}\subseteq \mathcal{Y}_{s}$ is the set of candidate labels associated with $\mathbf{x}_{i}$ and  $\begin{vmatrix} S_{i} \end{vmatrix}$ represents the number of candidate labels for instance $\mathbf{x}_{i}$. Particularly, instance $\mathbf{x}_{i}$ is annotated by a label vector $\mathbf{y}_{i}=\left\{ y_{c} \right\}^{Q}_{c=1}$, where $y_{i} \in\{0,1\}$ denotes whether the label $y_{c}$ is present in the candidate labels (`1') or not(`0'). When testing, CLIP-PZSL predict a set of unseen compositions $\mathcal{Y}_{u}$ that is mutually exclusive with training labels $\mathcal{Y}_{u}$: $ \mathcal{Y}_{s}\cap\mathcal{Y}_{u} = \oslash $. The goal is to remove the influence of noise and learn a classifier $\boldsymbol{f}:\mathcal{X}\mapsto  \mathcal{Y}_{s}\cup\mathcal{Y}_{u}$ that can predict true labels of unseen classes.

\subsection{Feature Extraction}

As shown in Fig.~\ref{model}, CLIP-PZSL first extracts instance and label features using CLIP, which contains an Image encoder and a Text encoder jointly trained on large-scale image–text pairs. For zero-shot classification, a label set is defined in natural language. Given an instance and set of labels, each label is embedded within a prompt to produce a “caption” is “A photo of a $\left\{ \right\}$.”. Let $\mathcal{Y}_{s}\cup\mathcal{Y}_{u}=\left\{  t_{c}:c=1,\ldots,K \right\}$ be the set of texts after such a prompt, where $K$ contains the number of seen and unseen classes. Then, the text $t_{i}$ is encoded by the CLIP Text encoder as $\mathbf{c}{i}=\boldsymbol{T}(t{i})\in\mathbb{R}^{d}$, where $\boldsymbol{T}(\cdot)$  denotes the CLIP Text encoder and $ \mathbf{c}_{i} \in\mathbb{R}^{d}$ with $d$ being the dimension of the label embedding. Meanwhile, each instance $\mathbf{x}_{i}$ is encoded by the Image encoder as $\mathbf{p}{i}=\boldsymbol{E}(\mathbf{x}_{i})\in\mathbb{R}^{d}$, where $\boldsymbol{E}(\cdot)$ denotes the Image encoder and $ \mathbf{p}_{i} \in\mathbb{R}^{d}$.

\begin{figure}
\centerline{\includegraphics[width=.49\textwidth]{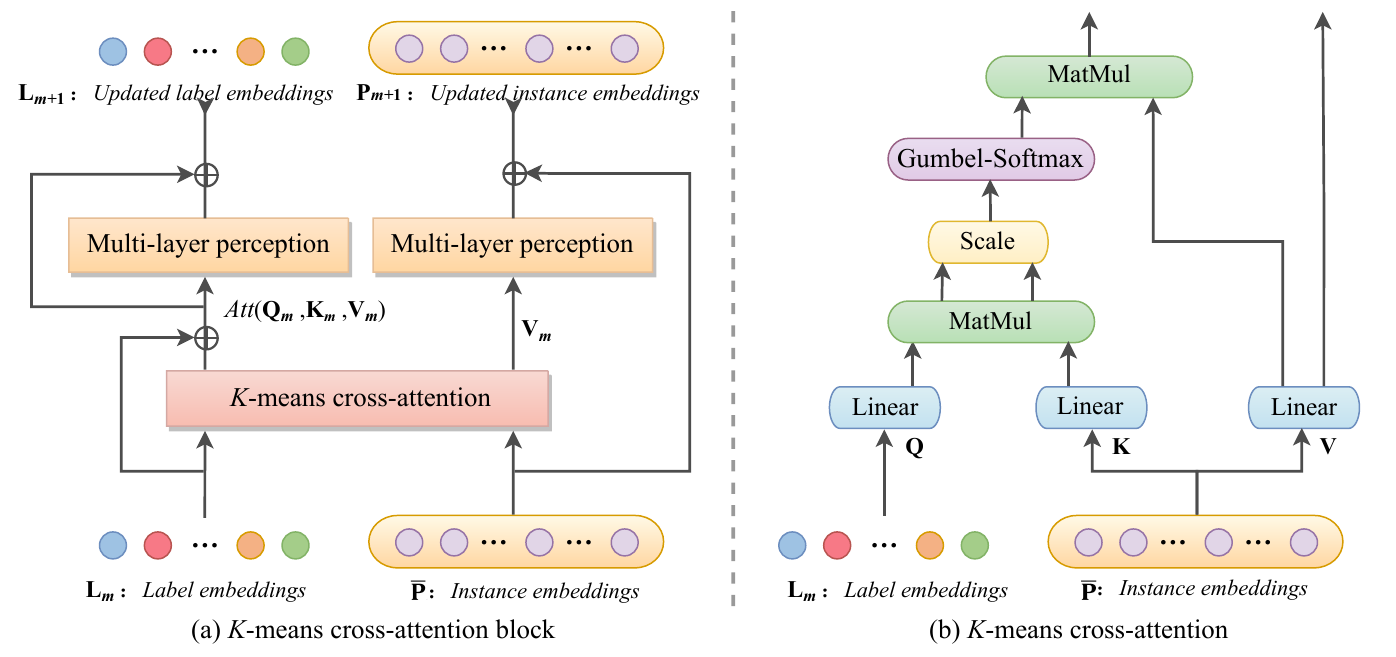}}
\vspace{-0.2cm}
\caption{Detailed structure of $K$-means cross-attention.}
\label{kmt}
\end{figure}

\subsection{Semantic Mining Block}

\textbf{Label Embeddings.} The label embeddings can be denoted as $\mathbf{L}=\left[ \mathbf{l}_{1},\mathbf{l}_{2},\ldots,\mathbf{l}_{K} \right] ^{T}\in\mathbb{R}^{K \times d}$, where each row is a label embedding. $\mathbf{L}$ is the learnable parameters during training and initialized with the text embedding from CLIP (i.e., $\mathbf{l}_{i}=\mathbf{c}_{i}$).

\textbf{Transformer Architecture.} We apply a new Transformer architecture to pool class-related features in CLIP instance embeddings, further exploring the potential feature relationship between instances and labels for semantic alignment, which consists of self-attention, $K$-means cross-attention \cite{F11} and multi-layer perception. Fig.~\ref{kmt} depicts the structure of $K$-means cross-attention. Its inputs are label and instance embeddings, enabling extraction of key instance information and matching with label embeddings to detect noisy labels.

Specifically, let $\mathbf{L}_{m}$ as the label embeddings at the $m$-th layer, where $m \in [0, M]$. $\mathbf{L}_{0}=\mathbf{L}$ and $M$ is the number of $K$-means cross-attention transformer layers. $\mathbf{P}=\left[ \mathbf{p}_{1},\ldots,\mathbf{p}_{i},\ldots, \mathbf{p}_{N}\right] ^{T} \in\mathbb{R}^{N \times d}$ is the instance embedding matrix from CLIP image encoder, and $\bar{\mathbf{P}}$ is its output after self-attention. Then, we use label embeddings of the linear transformation layer as query $\mathbf{Q}_{m}$ and perform $K$-means cross-attention to pool label features from the instance embeddings. Meanwhile, we respectively obtain key $\mathbf{K}_{m}$ and value $\mathbf{V}_{m}$ projections of $\bar{\mathbf{P}}$ by linear transformation layers:
\begin{equation}
	\begin{split} \mathbf{Q}_{m}=\mathbf{L}_{m} \mathbf{W}_{m}^{Q}, \quad \mathbf{K}_{m}=\bar{\mathbf{P}} \mathbf{W}_{m}^{K}, \quad \mathbf{V}_{m}=\bar{\mathbf{P}} \mathbf{W}_{m}^{V}
	\end{split}
\end{equation}  

\noindent where $\mathbf{W}_{m}^{Q},\mathbf{W}_{m}^{K},\mathbf{W}_{m}^{V}\in\mathbb{R}^{d\times d}$ are learnable parameter matrices of query, key, and value. The query and a set of key pairs to compute an attention map. Values are weighted and summed with the weight on the attention map to output the hidden feature of the label class for the following layer. Therefore, the $K$-means cross-attention can be calculated as:
\begin{equation}
	\begin{split} 
		Att(\mathbf{Q}_{m},\mathbf{K}_{m},\mathbf{V}_{m})=\omega\left(\frac{\mathbf{Q}_{m} \times \mathbf{K}_{m}^{T}}{\sqrt{d}}\right) \mathbf{V}_{m} 
	\end{split}
\end{equation}

\noindent where $\sqrt{d}$ is a scaling factor based on the depth of the network and $d$ is the feature dimension; $\omega(\cdot)$ is a Gumbel-Softmax \cite{jang2017categorical} function to approximate argmax function. This operation updates label embeddings to obtain better class-related features by taking a weighted average of the same label classes. 

\textbf{Feature Projection.} The output of the $K$-means cross-attention is then normalized and fed into the MLP to generate the input to the next block. Finally, we will get the label embeddings $\mathbf{L}_{M}\in \mathbb{R}^{K \times d}$ for $K$ classes and instance embeddings $\mathbf{P}_{M}\in \mathbb{R}^{N \times d}$. Then,  a linear classifier is used to classify the instance embeddings based on $\mathbf{P}_{M}$:
\begin{equation} 
	\begin{split} 
		\mathbf{M}=\operatorname{softmax}\left(\mathbf{P}_{M} \mathbf{W}+ \mathbf{b}\right)
	\end{split}
\end{equation} 

\noindent where $\mathbf{W}\in \mathbb{R}^{d \times Q}$ and $\mathbf{b}$ are the parameters to be learned. $\mathbf{M}$ is the predicted label confidence matrix, where only seen $Q$ classes in the training dataset are predicted.

\subsection{Instance-label Alignment with Partial Zero-shot Loss}

In this phase, the goal is to identify potential ground-truth label in candidate label set to reduce the impact of ambiguous labels. We compute cosine similarity between instance and text embedding to detect noisy labels in candidate label set: 
\begin{equation}
	\begin{split} 
		r_{ij}^{t} = \frac{\mathbf{p}_{i}^{t} \mathbf{c}_{j}^{T}}{\left\|\mathbf{p}_{i}^{t}\right\|_{2}\left\|\mathbf{c}_{j}^{T}\right\|_{2}}
	\end{split}
\end{equation} 

\noindent where $r_{ij}^{t}$ measures the similarity between instance $\mathbf{x}_{i}$ and text embedding $\mathbf{c}_{j}$, $\mathbf{c}_{j}$ corresponds to the label $y_{j}$ and $\mathbf{p}_{i}^{t}$ represents the embedding features of instance $\mathbf{x}_{i}$ in the $t$-th iteration. A larger $r_{ij}$ indicates the instance has a greater probability of being labeled $y_{j}$. According to Eq.(4), the label correction matrix $\mathbf{R}\in\mathbb{R}^{N \times Q}$ can be calculated, where $r_{ij} \in \mathbf{R}$.

The partial zero-shot loss is defined as
\begin{equation}
	\begin{split} 
        \mathcal{L}&=\mathcal{L}_{ce}(\mathbf{M},\mathbf{R},\mathbf{Y})+\mathcal{L}_{dist}(\mathbf{C},\mathbf{L})\\
		&=-\sum_{i=1}^{N}\sum_{j=1}^{Q} (r_{ij}^{t} \mathbf{Y}_{ij}^{t}\log \mathbf{M}_{ij}^{t})+\left\|\mathbf{C}-\mathbf{L}^{t}\right\|_{2}^{2} 
	\end{split}
\end{equation} 
\vspace{-0.2cm}

\noindent where $r_{ij}$ is correction weight, which can reduce the impact of noise labels by increasing the weight of instances and similar label embedding. $\mathbf{Y}_{ij}^{t}$ represents label confidence weight at the $t$-th iteration, i.e., the probability that $y_{j}$ is the ground-truth label of $\mathbf{x}_{i}$. $\mathbf{M}_{ij}^{t}$ is the predict value of $\mathbf{x}_{i}$ as $y_{j}$ in the $t$-th iteration. $\mathbf{C}$ is text embedding learned from CLIP and $\mathbf{L}^{t}$ represents the label embeddings learned in the $t$-th iteration. Confidence weights are iteratively refined:

\vspace{-0.2cm}
\begin{equation}
	\begin{split} 
		\mathbf{Y}_{ij}^{t+1}=\left\{\begin{array}{ll}
			\mathbf{U}_{ij}^{t} / \sum_{y_{k} \in S_{i}} \mathbf{U}_{ik}^{t}, & \text { if } y_{j} \in S_{i} \\
			0, & \text { otherwise }
		\end{array}\right.
	\end{split}
\end{equation} 

\noindent where $\mathbf{U}_{ij}^{t}=r_{ij}^{t}+\mathbf{M}_{ij}^{t}$. Therefore, Eq.(6) is intuitively reasonable. For the first part of cross entropy loss, as the training epochs grow, the ground-truth labels are identified incrementally, while refined labels in turn help to improve the classifier by guiding model learning. For the second part of mean square error loss, which can aligns instance and label embeddings in the same dimension to minimize semantic mismatch. 

Thus, for each test instance, $\tilde{y}$ denotes the predicted result of the instance $\mathbf{x}$ that has the highest label score:

\begin{equation} \label{8}
	\begin{array}{l}
		\tilde{y}=\underset{\mathcal{Y}_{s} \cup \mathcal{Y}_{u}}{\arg \max } (\mathbf{p}\mathbf{C}^{T})
	\end{array}
\end{equation}

\subsection{Computational Complexity} 
CLIP-PZSL consists of two core modules: semantic mining block and instance-label alignment block. In the semantic mining block, self-attention mechanism costs $\mathcal{O}(Nd^{2})$, with additional transformations in the $K$-means cross-attention block resulting in $\mathcal{O}(3Nd^{2})$. The computations for $\mathbf{Q}_{m} \times \mathbf{K}_{m}^{T}$ and Gumbel-Softmax both contribute $\mathcal{O}(NKd)$, while the MLP adds $\mathcal{O}(Nd^{2}+Kd^{2})$. Since $K\ll N$, the complexity simplifies to $\mathcal{O}(Nd^{2})$. In the instance-label alignment block, the label correction matrix requires $\mathcal{O}(NKd)$, and the loss functions contribute $\mathcal{O}(NK)$ and $\mathcal{O}(Nd)$. Thus, the overall complexity is $\mathcal{O}(Epoch(3Nd^{2}+Nd))$.

\section{Experiments}

\subsection{Experimental Setup}

\textbf{Datasets.} We select six public ZSL benchmarks: \textbf{CIFAR-10} \cite{F17} (10 classes, divided into 8 seen/2 unseen), \textbf{CIFAR-100} \cite{F17} (100 classes, 80 seen/20 unseen), \textbf{Food-101} \cite{F18} (101 classes, 80 seen/21 unseen), Caltech-UCSD Birds-200-2011 (\textbf{CUB}) \cite{F19} (200 classes, 150 seen/50 unseen), \textbf{Flowers-102} \cite{F20} (102 classes, 80 seen/22 unseen), Animals with Attributes 2 (\textbf{AWA2}) \cite{F21} (50 classes, 40 seen/10 unseen).

\textbf{Synthesized partial zero-shot datasets.} Following the PLL dataset synthesis method \cite{F22,F23}, we employ $Q-1$ independent decisions on noise labels to generate a candidate label set of instance, where $q \in {0.1, 0.3, 0.5}$ controls the probability of including a noise label. Higher $q$ implies more noise and greater disambiguation difficulty. Irrelevant labels are randomly selected as candidates across all benchmarks.

\textbf{Implementation Details.} We use the CLIP \cite{F12} with ViT-B/16 for Image and Text encoders. Images are resized to 224×224. CLIP-PZSL uses $M=3$ consecutive transformer layers, and is trained for $t=100$ epochs with the SGD optimizer (momentum=0.9), a mini-batch size of 64, and a learning rate of 0.001. The model is implemented in PyTorch.

\subsection{Compared Methods.}
We compare CLIP-PZSL with six state-of-the-art ZSL methods, including CLIP \cite{F12}, CALIP \cite{F24}, ABP \cite{F25}, SDGZSL \cite{F26}, Transzero \cite{chen2022transzero}, and CoAR-ZSL \cite{du2023boosting}. CLIP-based methods are evaluated on all datasets. While traditional methods (marked in gray) rely on attribute information, applicable only for the AWA2 and CUB datasets.

\subsection{Main Experiment Results and Analysis} 

Table~\ref{Tab01} presents results with the accuracy of seen classes (S.Acc) and unseen classes (U.Acc).

\begin{table}
	\scriptsize	
	\centering	
	\caption{The comparison with other ZSL methods.}
    \scalebox{1.0}{
		\label{Tab01}
			\begin{tabular}{cc|p{0.42cm}p{0.42cm}|p{0.42cm}p{0.42cm}|p{0.42cm}p{0.42cm}}		
				\hline\noalign{\smallskip}		
				\multirow{2}{*}{Dataset}&\multirow{2}{*}{Method} & \multicolumn{2}{c}{$q = 0.1$}& \multicolumn{2}{c}{$q = 0.3$}& \multicolumn{2}{c}{$q = 0.5$} \\		
				\cmidrule(r){3-8} 	
				&	&  S.Acc  & U.Acc   &  S.Acc  & U.Acc &  S.Acc  & U.Acc  \\		
				\midrule				
				&CLIP  &87.23   & 89.90 &87.23   & 89.90  &87.23   & 89.90    \\  
				\textbf{CIFAR-10}&CALIP  &85.30   & 93.25   &85.30   & 93.25    &85.30   & 93.25   \\  
				&CLIP-PZSL  &\textbf{92.15}  &\textbf{95.45}  &\textbf{91.97}  &\textbf{95.40} &\textbf{91.71} &\textbf{95.3}  \\  
				\midrule	
				&CLIP  &62.36   & 61.85   &62.36   & 61.85  &62.36   & 61.85 \\  
				\textbf{CIFAR-100}&CALIP  &61.71   & 63.25  &61.71   & 63.25   &61.71  & 63.25    \\  
				&CLIP-PZSL  &\textbf{74.83}  &\textbf{64.10}   &\textbf{74.30}  &\textbf{64.05}  &\textbf{71.21} &\textbf{64.01}  \\ 
				\midrule	
				&CLIP  &81.13  & 79.29   &81.13  & 79.29   &81.13  & 79.29  \\  
				\textbf{Food-101}&CALIP  &79.77   & 78.03  &79.77   & 78.03  &79.77   & 78.03  \\  
				&CLIP-PZSL  &\textbf{89.46}  &\textbf{80.24}  &\textbf{88.8}   &\textbf{80.17} &\textbf{87.58} &\textbf{80.07}  \\ 
				\midrule	
				&CLIP  &61.67  & 71.63   &61.67  & 71.63   &61.67  & 71.63  \\  
				\textbf{Flowers-102}&CALIP  &59.22  & 63.61   &59.22  & 63.61   &59.22  & 63.61   \\  
				&CLIP-PZSL  &\textbf{89.86}  &\textbf{71.82}  &\textbf{87.09}  &\textbf{71.77} &\textbf{80.44} &\textbf{71.73}   \\ 
				\midrule	
				&CLIP  &92.64  & 89.86   &92.64  & 89.86 &92.64  & 89.86  \\  
				&CALIP  &92.10  & 89.54  &92.10  & 89.54  &92.10  & 89.54  \\ 
                \rowcolor{gray!25}
				&{ABP}  &53.55  & 9.51   &52.73  & 8.47  & 9.64   & 5.49  \\ 
                \rowcolor{gray!25}
				\textbf{AWA2}&{SDGZSL}  &42.56  & 11.82 &42.41  & 10.23 &41.00 & 9.77   \\
                \rowcolor{gray!25}
                &{Transzero}  &30.94  & 56.51  &12.82  & 36.77 &3.33  & 9.65\\ 
                \rowcolor{gray!25}
                &{CoAR-ZSL}  &77.86  & 50.87 &41.74  & 36.53  &37.27  & 28.76  \\ 
				&CLIP-PZSL  &\textbf{95.09}  &\textbf{90.37}  &\textbf{95.08}  &\textbf{90.33}  &\textbf{94.52} &\textbf{90.32}  \\ 			
				\midrule	
				&CLIP  &46.57  & 41.96  &46.57  & 41.96  &46.57  & 41.96 \\  
				&CALIP  &43.41   & 35.21   &43.41   & 35.21  &43.41  & 35.21\\  
                \rowcolor{gray!25}
				&{ABP}  &33.46  & 2.16  &28.81  & 1.47   &28.66  & 1.27 \\ 
                \rowcolor{gray!25}
				\textbf{CUB}&{SDGZSL}  &15.42  & 1.95   &14.35  & 2.00  &14.05  & 1.84  \\ 
                \rowcolor{gray!25}
                &{Transzero}  &22.62  & 8.43  &0.88  & 2.47 &0.88  & 1.16    \\ 
                \rowcolor{gray!25}
                &{CoAR-ZSL}  &17.40  & 16.82  &2.95  & 3.37  &1.36  & 1.92  \\ 
				&CLIP-PZSL  &\textbf{58.27}  &\textbf{42.16} &\textbf{51.04}  &\textbf{42.16}  &\textbf{47.68} &\textbf{42.16} \\ 
				\noalign{\smallskip}\hline
			\end{tabular}
    }
\end{table}

\textbf{CLIP-based ZSL methods.} CLIP-PZSL shows superior performance on all datasets compared to the CLIP-based ZSL methods, with significant improvements in both S.Acc and U.Acc. This indicates that compared with pre-trained ZSL models, CLIP-PZSL can better reduce the impact of ambiguous labels and achieve effective alignment of instance and label embeddings, thereby improving the generalization ability of the model in unseen classes.

\textbf{Traditional ZSL methods.} Experiments on the \textbf{AWA2} and \textbf{CUB} datasets show that traditional methods degrade markedly in the presence of ambiguous labels, as they tend to overfit noisy annotations. This overfitting limits the model's ability to learn effective, meaningful information and generalize to unseen classes, leading to a drop in overall performance. In contrast, CLIP-PZSL retains a clear advantage even under such challenging conditions, outperforming traditional methods across all scenarios.

\begin{table}[!htbp]
  \centering
  \vspace{-0.2cm}
  \caption{Ablation study on Food-101 and CUB with $q=0.3$.}
  \label{tab:abl}
  \scalebox{0.9}{
  \begin{tabular}{*{5}{c}}
    \toprule
    \multirow{2}{*}{Ablation} & \multicolumn{2}{c}{CUB} & \multicolumn{2}{c}{Food-101}\\
    \cmidrule(lr){2-3} \cmidrule(lr){4-5} 
    & S.Acc & U.Acc  & S.Acc & U.Acc\\
    \midrule
    w/o cross-entropy loss & 84.00 & 80.02 & 48.45 & 41.96 \\ 
    w/o mean square error loss & 82.05 & 80.04 & 45.81 & 41.68 \\ 
    w/o semantic mining block & 84.26 & 80.01 & 48.62 & 42.10 \\ 
    CLIP-PZSL & \textbf{88.80} & \textbf{80.17} & \textbf{51.04} & \textbf{42.16} \\ 
    \bottomrule
  \end{tabular}
  }
\end{table}

\subsection{Ablation study} 

Table~\ref{tab:abl} validated the importance of components in CLIP-PZSL, including the semantic mining block and partial zero-shot loss. Compared with w/o semantic mining block, CLIP-PZSL achieves stronger disambiguation and recognition on unseen classes, as the block extracts latent semantic information and reduces the effect of noisy labels. CLIP-PZSL also outperforms two variants involving loss, as $\mathcal{L}_{ce}$ learns weights for candidate labels, while $\mathcal{L}_{dist}$ aligns instance and label embeddings to reduce semantic mismatch. These two terms complement each other, further boosting performance.

\section{Conclusion} 

This paper proposes CLIP-PZSL to address ambiguous labels. The method introduces a semantic mining block, which integrates both instance and label semantic information to adaptively extract distinct features, allowing for the detection of noisy labels by assessing the similarity between instances and ambiguous labels. It improves the semantic alignment of embeddings in high-dimensional space. Additionally, the paper presents a new partial zero-shot loss, which not only reduces the impact of noisy labels but also aligns instance and label embeddings within the same dimension to minimize semantic discrepancies. As training progresses, the true labels are gradually identified, and these refined labels contribute to improving the classifier's performance on unseen classes. Experiments verify CLIP-PZSL's effectiveness.

\clearpage

\label{sec:refs}

\small 

\bibliographystyle{IEEEbib}
\bibliography{refs}

\end{document}